\documentclass[sigconf,nonacm]{acmart}
\usepackage{graphicx}
\usepackage{amsmath}
\usepackage{subfigure}
\usepackage{bm}
\usepackage{bbm}
\usepackage{algorithm}
\usepackage{algpascal}
\usepackage{algpseudocode}

\AtBeginDocument{%
  \providecommand\BibTeX{{%
    \normalfont B\kern-0.5em{\scshape i\kern-0.25em b}\kern-0.8em\TeX}}}

\setcopyright{none}

\begin{document}

\title{Fast top-K Cosine Similarity Search through XOR-Friendly Binary Quantization on GPUs}

\author{Xiaozheng Jian}
\affiliation{%
  \institution{Tencent}
}
\email{ajian@tencent.com}

\author{Jianqiu Lu}
\affiliation{%
  \institution{Tencent}
}
\email{jianqiulu@tencent.com}

\author{Zexi Yuan}
\affiliation{%
  \institution{Tencent}
}
\email{percyyuan@tencent.com}

\author{Ao Li}
\affiliation{%
  \institution{Tencent}
}
\email{aoli@tencent.com}


\begin{abstract}
  We explore the use of GPU for accelerating large scale nearest neighbor search and we propose a fast vector-quantization-based exhaustive nearest neighbor search algorithm that can achieve high accuracy without any indexing construction specifically designed for cosine similarity. This algorithm uses a novel XOR-friendly binary quantization method to encode floating-point numbers such that high-complexity multiplications can be optimized as low-complexity bitwise operations. 
Experiments show that, our quantization method takes short preprocessing time, and helps make the search speed of our exhaustive search method much more faster than that of popular approximate nearest neighbor algorithms when high accuracy is needed.
\end{abstract}




\maketitle

\section{Introduction}
It is hard to find specific content in massive resource library. Generally, these contents can be transformed into vectors of different lengths using proper embedding algorithms. The state-of-the-art examples include, for text data, word2vec \cite{mikolov2013distributed}, and for image data, convolutional neural network \cite{sharif2014cnn, gong2014multi}.
Using these embedding vectors of dozens to thousands of dimensions, the distance from queries to every entry in the database can be calculated, and the nearest ones can be found. The true problem is how to find the most similar contents from an arbitrary query in a large database when users request it with low delay. This is also the most computationally expensive part of many algorithms diversified in biology (gene classification \cite{pan2004comprehensive}), computer vision (local image feature matching \cite{lowe2004distinctive}), speech recognition (content-based music search \cite{li2004content}) and many other fields. 

We hope that the both speed and accuracy of the search algorithms can be high even employed on large datasets. When working with high-dimensional features, which is often the case in computer vision applications, there is no known exact nearest-neighbor search algorithm that has acceptable performance. To obtain a speed improvement, researchers developed approximate search algorithms \cite{aumuller2017ann}. Generally these algorithms can provide 80 percent or more of the correct neighbors, and be much faster than exact search. When a even higher proportion of correctness is required, for example 90 percent or 98 percent, the speed of most approximate search algorithms drops quickly. Furthermore, Some of these algorithms need to train a codebook for indexing before searching, which is also a time-consuming part.

Since the data size is becoming extremely large and the embedding vectors can be long to keep more information, the distance calculation now requires a great number of basic arithmetic calculations and comparisons, but the calculation for pairs of embedding vectors does not affect each other. In this decade, great development on GPU computation has been achieved. GPUs are much skilled at parallel computing for simple calculation than CPUs, which meets the case of nearest-neighbor search problem. Exact search algorithm \cite{garcia2010k} and some approximate search algorithms \cite{pan2011fast,johnson2019billion} have already been implemented on GPU and obtained great improvements on performance.

In this paper, we propose a fast and accurate algorithm implemented on GPU for approximate nearest-neighbor search problem. A binary quantization method is proposed to compress floating-point numbers into 3- or 4-bit binary codes without training. Cosine similarity calculations of vectors are simplified to exclusive OR (XOR) operations of binary codes and is further optimized based on the parallel characteristics of GPU. Based on the quantization method and optimized calculations, cosine similarities of normalized data points can then be fast calculated on GPU when both data size and vector length are large. As training is not needed for quantization, this method will be useful for the situations where dataset distribution changes rapidly or only few queries are asked in large datasets.

This paper makes the following contributions:

\begin{itemize}
  \item We propose a new quantization method to encode number within $[-1,1]$ into arbitrary bits (Section 3).
  \item We provide a novel view that transforming multiplication into bitwise XOR operation, and use this transformation to accelerate multiplications in limited scope (Section 3).
  \item We propose a train-free algorithm to implement GPU exhaustive kNN-Selection on large datasets, which based on cosine similarity and has a series of parameters controlling the accuracy and speed (Section 3 \& 4).
  \item We conduct real-data experiments that show that the proposed algorithm has a state-of-the-art searching efficiency and high accuracy on large-scale nearest-neighbor search tasks. The algorithm is also extensible on multi-GPU configurations (Section 5).
\end{itemize}


\section{Related Work}

Generally, K-nearest neighbor search is to find the top $K$ most similar vectors in $n$ vectors for each vector query given the distance metric. Here each vector has $N$ components, and in this paper we specify the metric with descending cosine similarity, defined by the inner product of two normalized vectors. This section presents a review of previous work in this area.

The brute force way to solve the problem is to calculate pairwise distance between the query vector and each alternative vector and use a minimum heap to store the top $K$ nearest vector. This way costs great computing resources (with time complexity $O(nN+n\log(K))$). Garcia et al. \cite{garcia2010k} implemented parallel brute force algorithm on NVIDIA GPU using CUDA and CUBLAS, showing that the speed can be 25x and 62x faster than highly optimized ANN C++ library implemented on CPU. As this migration cannot really reduce the resource consumption on computation, researchers have been providing solutions to calculate the approximate nearest neighbors with high precision but much lower time complexity. Most of these techniques are target on reduce the search space. We revise the most widely used K-NN search techniques, classified in three categories: hashing based techniques, partitioning trees and nearest neighbor graph techniques.

The best-known hashing-based techniques might be local sensitive hashing (LSH) \cite{andoni2006near}, in which many hash functions, with property that elements with similar hashes are more likely to be similar, are used. Variants of LSH such as multi-probe LSH \cite{lv2007multi} and LSH Forest \cite{bawa2005lsh} help improve the performance of these techniques. As the performance of LSH is highly related to the quality of hashing functions, huge work focuses on improving hashing methods \cite{shakhnarovich2003fast,wang2010semi,xu2011complementary}. Pan implemented LSH based nearest-neighbor search on GPU \cite{pan2011fast}, making searching much faster. As LSH is highly sensitive to the hashing function we choose, we are not going to compare it with our method in the experiment section. However, combining LSH with our method to further accelerate the search speed is possible, as we will mention later.

Partitioning trees is also a popular technique for approximate nearest-neighbor search. KD tree \cite{silpa2008optimised} is one of the best known nearest-neighbor algorithms. It is effective on datasets with low dimensionality but gets poor performance on datasets with high dimensionality. Gong encoded image matrix into binary values to find similar results and achieved good search recalls for image datasets \cite{gong2012angular,gong2013learning}. This method does not provide good results for non-classification problems. Other methods like annoy \cite{bernhardsson2018annoy}, ball tree \cite{liu2006new,omohundro1989five} use decision tree to make searching an $O(\log(N))$ level job.

Herve Jegou proposed product quantization (PQ) to provide a short code representation of vectors \cite{jegou2010product} and improved the searching efficiency by IVFADC algorithm \cite{jegou2011searching}. In product quantization, space is decomposed into a Cartesian product of low dimensional subspaces, and data points are represented by compact codes computed as quantization indices in these subspaces. A codebook needs to be trained by a training dataset with distribution similar to the population before indexing. The training phase can take a long time when the training set is large. The compact codes are then efficiently compared to the query points using an asymmetric approximate distance in the search phase. Using an inverted file system, PQ can help efficiently search nearest neighbors on high-dimensional datasets. Inverted multi-index (IMI) proposed by Babenko and Lempitsky \cite{babenko2014inverted}, which replaces the standard quantization in an inverted index with product quantization, obtains a denser subdivision of the search space. These methods are efficient at searching on large datasets with high dimensionally and are now used and accelerated by GPU in Facebook's Faiss library \cite{johnson2019billion}. As the new approach proposed in this paper is also a vector-quantization based technique, we will compare our results with the IVFADC version of PQ based nearest-neighbor searching. 

Nearest neighbor graph methods is based on the thought that, when there comes a query, we start to calculate the distance from a random point to the query, and try to search along the "steepest descent direction" of distance between points on the direction and the query. In practice, a graph structure in which points are vertices and edges connect each point to its friend points is built. For each point in the graph, the friend points are likely to be close to it. The query points are used to explore this graph using various strategies in order to get closer to their nearest neighbors. As an optimized graph must be built, these graph methods also have train phases to build graph and take long time on training when the dataset is huge. Malkov raised an efficient and robust searching algorithm using Hierarchical Navigable Small World (HNSW) graphs \cite{malkov2018efficient}. HNSW is one of the best practices of nearest neighbor graph techniques so far. However, it is not a good idea to implement HNSW on GPU, which takes advantage on parallel computing, since we need to access the points based on a strict hierarchical order. We will compare the performance of our approach and HNSW with comparable computing resources.  

Faiss \cite{johnson2019billion} is a good solution that works on GPU verified by ANN-benchmark \cite{aumuller2017ann}. The performance of PQ methods on GPU has been optimized by Faiss. The library also integrates a CPU version of HNSW search. For both algorithms here, the training and indexing step for large dataset takes a long time, which is a crucial drawback for some real-time online data. They suffer from a painful trade-off among training time, recall/precision rate and search speed. In the following section we will provide a novel approach with short preprocessing time, high recall/precision and fast search speed.


\section{Compress Vectors with XOR-friendly Binary Quantization}
Given two floating-point vectors $ \bm{X} = (x_1, x_2, \cdots, x_N) $ and $ \bm{Y} = (y_1, y_2, \cdots, y_N) $ with $\Vert\bm{X}\Vert_2=\Vert\bm{Y}\Vert_2=1$, the cosine similarity of them is 
$$
    \text{Similarity}(\bm{X},\bm{Y}) = \sum\limits_{i=1}^{N} x_i \cdot y_i 
$$
It requires N multiplications and (N-1) additions, resulting in intensive computational complexity though these operations can be parallelized by SIMD instructions. Besides, for floating-point vectors, the memory bandwidth can also limit the throughput of computations when processing large scale similarity computation. For example, the memory bandwidth of DDR4 2666 is 21.3 GB/s. To solve these two problems, we propose a fast similarity search mechanism that quantizes 32-bit floating-point numbers to low-bit binary numbers and replaces high-complexity multiplications by XOR computations, which has low complexity on computation. We first introduce the relationship between multiplication and XOR operation.

\subsection{Multiplication and XOR on simple sets}

\par Consider two sets $G=\{+1, -1\}$ and $\bar{G}=\{0,1\}$. Define multiplication on $G$ is simple multiplication ($\cdot$), and multiplication on $\bar{G}$ is XOR operation ($\oplus$). The following proposition reveals the relationship between these two different operations.

\begin{proposition} \label{prop:groups}
$(G, \cdot)$ and $(\bar{G}, \oplus)$ are two isomorphism groups under mapping $\sigma: G\rightarrow\bar{G}$, where
\begin{equation} \label{eq:prop_1}
\sigma(a)=\frac{1-a}{2}, \forall a\in G
\end{equation}
\end{proposition}
The proposition can be directly verified by checking enumerate all possible operations. In these two groups, (+1) and 0 are identity elements while (-1) and 1 are inverse elements respectively. 

Using \ref{prop:groups}, the multiplication on $G$ can be replaced by the XOR computation on $\bar{G}$ with the following formula.
\begin{equation} \label{eq:mul_xor:2}
    \begin{aligned}
        a_1 \cdot a_2 &= \sigma^{-1}{({\bar{a}_1 \oplus \bar{a}_2})} \\
                &= 1 - 2({\bar{a}_1 \oplus \bar{a}_2})
    \end{aligned}
\end{equation}
Therefore, if all components in vectors can be represented as a combination of $ a_i \in \{+1, -1\}$, then the high-complexity multiplication can be replaced by the low-complexity XOR computation.

\subsection{XOR-Friendly Binary Quantization}
To take advantage of the XOR computation, a special strategy is used to quantize floating-point numbers to binary numbers. 

For any real number $x\in(-1,+1)$, we use a mapping $f_B(\cdot)$ $(B\geq 1)$ to approximate $x$ into a subspace of $\mathbb{R}$. For simplification we write $x_B=f_B(x)$. $f_B$ has the following representation:
$$ \label{eq:product:0}
    f_B(x)=x_B = a_{B-1} \cdot \frac{1}{2} + \cdots + a_{1} \cdot \frac{1}{2^{B-1}} + a_0 \cdot \frac{1}{2^B} = \sum\limits_{i=0}^{B-1} \frac{1}{2^{B-i}} a_i
$$
where $ a_i \in G,\ i\in{0,1,...,B-1} $. Specially we define $f_0(x)=0$. Then the value of $a_i$ is decided by the following formula:
$$ \label{eq:product:how}
  a_{B-1-i} = sign(x-x_i\geq 0)\,\ i\in{0,1,...,B-1}, 
  sign(x)=
\left\{ \begin{aligned} 
1 & & x\geq0 \\ 
-1 & & x<0 \\ 
\end{aligned} \right .
$$

Similar with binary representation, this approximation has some good properties. Define $\mathbb{N}^+$ as the set of all positive integer, then we have
\begin{proposition} \label{prop:limit}
$\forall x \in(-1,+1),\ \forall B\in \mathbb{N}^+,\ |x_B-x|\leq2^{-B}$. And $x_B$ uniformly convergent to $I(x)=x$: 
$\lim_{B\rightarrow+\infty} x_B = x.$
\end{proposition}
The proof of this proposition is in Appendix \ref{sec:prof1}.

When $ B \rightarrow +\infty $, $ x $ can be exactly represented. Instead, if $ B $ is fixed to a finite number, $ x $ will be approximately represented. In this case, if we take a fixed $B$ (indicating the number of encoding bits), with the vector $\textbf{C} = [\frac{1}{2}, \frac{1}{2^2}, \cdots,  \frac{1}{2^B}] $ as a fixed codebook, denote $[a_{B-1}, \cdots, a_1, a_0]\cdot\textbf{C}$ as $(a_{B-1} \cdots a_1 a_0)_{(\cdot)}$, then all $x \in (-1, +1)$ can then be encoded as
\begin{equation} \label{eq:product:1}
    x = (a_{B-1} \cdots a_1 a_0)_{(\cdot)}  
\end{equation}
where $ a_i \in G $. 

Next we will show that, based on this quantization scheme, the multiplication between floating-point numbers can then be replaced by XOR. Given the encoding bit $B_x, B_y$, the product of $x,y \in (-1, +1)$ can be calculated as
\begin{equation} \label{eq:product:2}
    \begin{aligned}
        xy &= (a_{B_x-1} \cdots a_1 a_0)_{(\cdot)} (c_{B_y-1} \cdots c_1 c_0)_{(\cdot)} \\
        &= \sum\limits_{i=0}^{B_x-1} \frac{1}{2^{B_x-i}} a_i \cdot \sum\limits_{i=0}^{B_y-1} \frac{1}{2^{B_y-i}} c_i 
        = \frac{1}{2^{B_x+B_y}} \sum\limits_{i=0}^{B_x-1}\sum\limits_{j=0}^{B_y-1} 2^{i+j} (a_ic_j) \\
    \end{aligned}
\end{equation}
where $a_i,c_i \in \{+1, -1\}$. By Eq. \ref{eq:mul_xor:2}, Eq. \ref{eq:product:2} can be transformed as
$$ \label{eq:product:3}
    \begin{aligned}
        xy &= \frac{1}{2^{B_x+B_y}} (\sum\limits_{i=0}^{B_x-1}\sum\limits_{j=0}^{B_y-1} 2^{i+j}\sigma^{-1}(\bar{a}_i \oplus \bar{c}_j))\\
        &= \frac{1}{2^{B_x+B_y}} (\sum\limits_{i=0}^{B_x-1}\sum\limits_{j=0}^{B_y-1} (2^{i+j}-(\bar{a}_i \oplus \bar{c}_j) << (i+j+1)))
    \end{aligned}
$$
where $\bar{a}_i,\bar{c}_i \in \{0, 1\}$ are the corresponding element in $G$ described by proposition \ref{prop:groups}, and $<<$ is shift logical left instruction ($a<<b=a*2^b$). Then, the multiplications are replaced by fast bitwise operations - XOR and bit shifts. For convenience of description, a new operator $\otimes$ is introduced. We define $\bar{x} = (\bar{a}_{B_x-1} \cdots \bar{a}_1 \bar{a}_0)_2$ and $ \bar{y} =  (\bar{c}_{B_y-1} \cdots \bar{c}_1 \bar{c}_0)_2 $. Here $\bar{x}$ and $\bar{y}$ are integers represented by binary code. We name this kind of quantization as XOR-Friendly Binary Quantization (XFBQ). We introduce a new operation $\otimes$ to denote 
\begin{align*}
 \label{eq:product:4}
        \bar{x} \otimes \bar{y} &= (\bar{a}_{B_x-1} \cdots \bar{a}_1 \bar{a}_0)_2 \otimes (\bar{c}_{B_y-1} \cdots \bar{c}_1 \bar{c}_0)_2 \\
        &= \sum\limits_{i=0}^{B_x-1}\sum\limits_{j=0}^{B_y-1} ((\bar{a}_i \oplus \bar{c}_j) \ll (i+j))        
\end{align*}
Then the result of $\bar{x} \otimes \bar{y}$ can be equivalently transformed to the product of $xy$
\begin{equation} \label{eq:product:6}
    \begin{aligned}
        xy &= \frac{1}{2^{B_x+B_y}} (\sum\limits_{i=0}^{B_x-1}\sum\limits_{j=0}^{B_y-1} 2^{i+j} - 2 \sum\limits_{i=0}^{B_x-1}\sum\limits_{j=0}^{B_y-1} (\bar{a}_i \oplus \bar{c}_j \ll (i+j))) \\
        &= \frac{1}{2^{B_x+B_y}} ((2^{B_x} - 1)(2^{B_y}-1) - 2 (\bar{x} \otimes \bar{y}))
    \end{aligned}
\end{equation}
Therefore, the multiplication of floating-point numbers can be optimized by following steps:
\begin{itemize}
    \item Given the encoding bit $B_x, B_y$, based on Equation \ref{eq:product:1} and \ref{eq:prop_1}, quantize two operands $x, y \in (-1, +1)$ as $\bar{x} =  (\bar{a}_{B_x-1} \cdots \bar{a}_1 \bar{a}_0)_2 $ and $\bar{y} =  (\bar{c}_{B_y-1} \cdots \bar{c}_1 \bar{c}_0)_2 $.
    \item Calculate the result of $ \bar{x} \otimes \bar{y}$ using XOR operations.
    \item Use Equation \ref{eq:product:6} to calculate the product of $xy$.
\end{itemize}

\subsection{Inner Product of Vectors with XFBQ}
Based the above quantization scheme of floating-point numbers, floating-point vectors can be quantized in a XOR-Friendly way using the following scheme. Given two N-dimension vectors $ \bm{X} = [x_0, x_1, \cdots, x_{N-1}]$ and $ \bm{Y} = [y_0, y_1, \cdots, y_{N-1}]$, $ x_k, y_k \in (-1, +1)$, with the encoding bit $B$, we have
\begin{equation} \label{eq:inner_product:0}
    \begin{aligned}
        x_k = (a_{(B_x-1)k} \cdots a_{1k} a_{0k})_{(\cdot)} &\Leftrightarrow \bar{x}_k = (\bar{a}_{(B_x-1)k} \cdots \bar{a}_{1k} \bar{a}_{0k})_2 \\
        y_k = (c_{(B_y-1)k} \cdots c_{1k} c_{0k})_{(\cdot)} &\Leftrightarrow \bar{y}_k = (\bar{c}_{(B_y-1)k} \cdots \bar{c}_{1k} \bar{c}_{0k})_2
    \end{aligned}
\end{equation}
where $a_{bk},c_{bk} \in \{+1, -1\}$ and $\bar{a}_{bk},\bar{c}_{bk} \in \{0, 1\}$. 
\par We denote $\sum_{k=0}^{N-1} \bar{x}_k \otimes \bar{y}_k$ with $\bar{\bm{X}} \otimes \bar{\bm{Y}}$. Using Eq. \ref{eq:product:6}, the result of $\bar{\bm{X}} \otimes \bar{\bm{Y}}$ can be transformed to the inner product by 
\begin{equation} \label{eq:inner_product:2}
    \bm{XY} = \frac{1}{2^{B_x+B_y}}(N(2^{B_x} - 1)(2^{B_y}-1) - 2(\bar{\bm{X}} \otimes \bar{\bm{Y}}))
\end{equation}
Furthermore, population count (POPCNT) operations are introduced to improve the performance of the inner product. Since
\begin{equation} \label{eq:inner_product:3}
    \begin{aligned}
        \bar{\bm{X}} \otimes \bar{\bm{Y}} &= \sum\limits_{k=0}^{N-1} \sum\limits_{i=0}^{B_x-1}\sum\limits_{j=0}^{B_y-1}((\bar{a}_{ik} \oplus \bar{c}_{jk}) << (i+j)), \\
    \end{aligned}
\end{equation}
notice that
\begin{align*}
 \label{eq:inner_product:4}
    \sum\limits_{k=0}^{N-1} (\bar{a}_{ik} \oplus \bar{c}_{jk}) &= POPCNT((\bar{a}_{i(N-1)} \cdots \bar{a}_{i0} )_2 .
    \oplus (\bar{c}_{j(N-1)} \cdots \bar{c}_{j0})_2)
\end{align*}
In order to utilize POPCNT operations and deduce the total number of instructions for a faster calculation performance, the N-dimension quantized vectors are considered to be reconstructed as $B_x(B_y)$-dimension vectors,
\begin{equation} \label{eq:inner_product:5}
    \begin{aligned}
        \hat{\bm{X}} = [\hat{\bm{x}}_{B_x-1}, \cdots, \hat{\bm{x}}_1, \hat{\bm{x}}_0],&\text{  where  }\hat{\bm{x}}_b = (\bar{a}_{b(N-1)} \cdots \bar{a}_{b0})_2 \\
        \hat{\bm{Y}} = [\hat{\bm{y}}_{B_y-1}, \cdots, \hat{\bm{y}}_1, \hat{\bm{y}}_0],&\text{  where  }\hat{\bm{y}}_b = (\bar{c}_{b(N-1)} \cdots \bar{c}_{b0})_2 \\
    \end{aligned}
\end{equation}

\begin{figure}[!htbp]
    \begin{center}
        \includegraphics[width=0.5\textwidth]{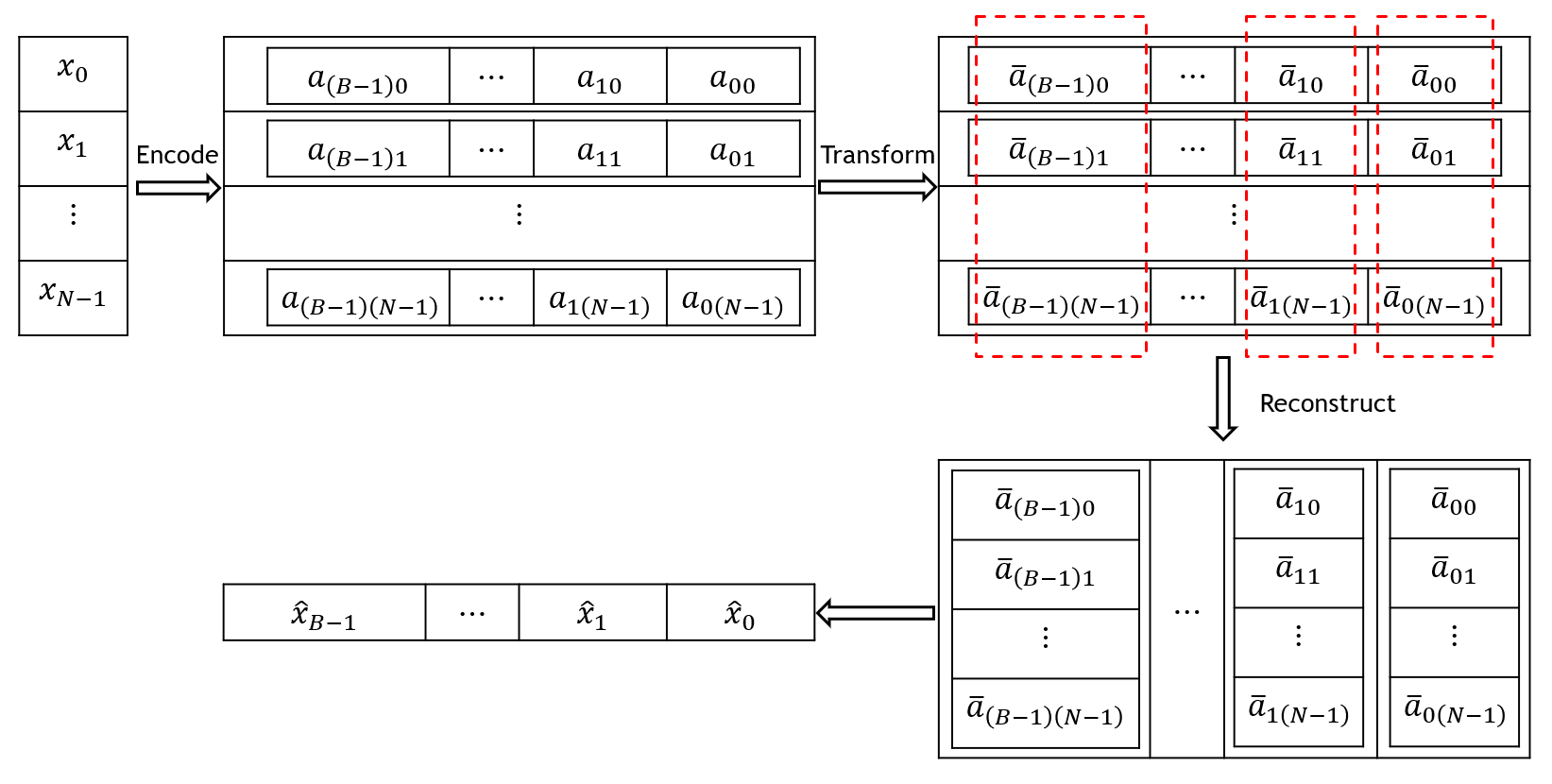}
    \end{center}
    \caption[]{Preprocessing of dataset in our method. The work will be done in time of O(BN). Encoding, transform and reconstruct can all be well done parallelly.}
    \label{fig:inner_product:0}
\end{figure}

The process of the reconstruction is shown in Figure \ref{fig:inner_product:0}. Then Eq. \ref{eq:inner_product:3} can be represented as
\begin{equation}\label{eq:inner_product:6}
        \hat{\bm{X}} \times \hat{\bm{Y}} \triangleq \sum\limits_{i=0}^{B_x-1}\sum\limits_{j=0}^{B_y-1} (POPCNT(\hat{\bm{x}}_i \oplus \hat{\bm{y}}_j) << (i+j)) 
        =  \bar{\bm{X}} \otimes \bar{\bm{Y}}.
\end{equation}

The benefit from this storage scheme can be shown by a example. Suppose $\bm{x}$ and $\bm{y}$ are both 32-d vectors, and we set $B_x=B_y=3$. Without encoding 32 multiplications are needed. IN XFBQ, both vectors are encoded into 96 bits. Without this scheme we need to take $96*96=9216$ XOR operations separately, which takes long time on sending instructions. If we store the data as $\hat{\bm{X}}$ and $\hat{\bm{Y}}$, then all $\hat{\bm{x}}_b$ and $\hat{\bm{y}}_b$ can be saved in 4-byte integers respectively as all bits in the same integer share the same shifting bits in calculation. Now both $\bm{x}$ and $\bm{y}$ are encoded into 3 integers, and only $3*3=9$ XOR and POPCNT operations with corresponding shifts are needed. As POPCNT is faster than multiplication, and we decrease the number of instructions, we can see great speedup. Modern processing units also support POPCNT64, which takes the same operation on two 8-byte integers. This instruction will further improve the performance. Based on this storage scheme, XFBQ works better on higher-dimensional vectors.

To sum all above up, calculating the inner product of two floating-point vectors can be optimized by following steps:
\begin{itemize}
    \item Given the encoding bit $B$, based on Equation \ref{eq:inner_product:0} and \ref{eq:inner_product:5}, quantize and reconstruct the vectors as $\hat{\bm{X}}$ and $\hat{\bm{Y}}$.
    \item Based on Equation \ref{eq:inner_product:6}, calculate the result of $\hat{\bm{X}} \times \hat{\bm{Y}}$.
    \item Based on Equation \ref{eq:inner_product:2} and \ref{eq:inner_product:6}, obtain the inner product of $\bm{X}\bm{Y}$.
\end{itemize}

\subsection{Complexity Analysis}
\label{subsec:complexity}
Given two N-dimension vectors $\bm{X}$ and $\bm{Y}$, here N is multiple of 64 for convenience of description, based on the above quantization scheme, these vectors can be quantized as $\hat{\bm{X}}$ and $\hat{\bm{Y}}$ and then perform bitwise operations instead of multiplications for the inner product. As shown in Table \ref{tab:complexity:0}, the calculation of the inner product can be significantly optimized by the proposed scheme.
\begin{table}[h]
    \caption{Analysis of computational complexity}
    \centering
    \begin{tabular} {c  c  c}
        \hline
        Calculation & Manipulations (times) & Memory (bytes) \\
        \hline \hline
        $\bm{XY}$ & \parbox{3cm}{\centering N multiplications and (N-1) additions} & 2N * size of(float) \\ [2ex]
        \hline 
        \parbox{1.5cm}{\centering $\hat{\bm{X}} \times \hat{\bm{Y}}$ }
        & \parbox{3cm}{\centering $B_xB_y * (N / 64)$ XOR and POPCNT, $(B_xB_y * (N / 64) - 1)$ bitwise shifts and additions }
        & \parbox{2cm}{\centering $(B_x+B_y)(N / 64)$ * size of(uint64)} \\ [25pt]
        \hline
    \end{tabular}
    \label{tab:complexity:0}
\end{table}

\subsection{Error Analysis of Inner Product Calculation}
\label{sec:err}
Using the quantization method we provided above, a 4-byte floating-point number can be easily transformed into and stored in 3 or 4 bits, saying that we can store 8-11 times data as storing the original data. We admit there is a huge loss on precision, but using the following tricks, the loss on the final precision of similarities will be acceptable. 

\par Suppose $\bm{x}, \bm{y}$ are two vectors with same dimension. $\bm{x'}$ and $\bm{y'}$ are approximations of them. Then Similarities is calculated by
\begin{equation}\label{calcsim}
\begin{aligned}
|\bm{x}||\bm{y}|\cos(\bm{x},\bm{y})&=sim(\bm{x},\bm{y})\\
&\approx sim(\bm{x'},\bm{y'})=\bm{x'}\cdot \bm{y'}=|\bm{x'}||\bm{y'}|\cos(\bm{x'},\bm{y'}).
\end{aligned}
\end{equation}
XFBQ can be seen as an approximate method. Using this formula, the similarity error introduced by quantization can be split into two parts: the length error and the angle error. 

\subsubsection{Length Error} For each component of a vector, it has been approximated to the nearest point that can be represented in a few bits, which introduce a length error in each dimension. As the error accumulates, the length of the quantized vector we stored should be different from the real value. That is the length error. Fixing the number of bits we use for one component, this error can be large when most of the components are small, which is just the case when vector has a lot of components. Good news is, in many fields like speech recognition and NLP, the data vectors are in high dimension, and we can expect that the vectors are usually small in every dimension with high probability. On the popular dataset of embedding vectors like GloVe \cite{pennington2014glove}, all tokens are no larger than 0.5 after vector normalization. Tokens are even smaller for many fields needing nearest neighbor search such as fingerprints recognition and facial recognition. These findings make the following assumption reasonable:

\newtheorem{assumption}{Assumption}[section]
\begin{assumption} 
$\forall \bm{x}=(x_1,...,x_N)$, $\exists \epsilon\in(0, 1/2]$, $s.t.$\\ $\max_{i\in\{1,...,N\}} x_i<\epsilon$.
\end{assumption}

Even for the cases that some vectors have large value in specific dimensions, we can take a linear transformation using an orthogonal matrix $R$ \cite{gong2012angular} to make the assumption effective. Generally, $\epsilon$ will shrink when $N$ goes high. We can expect that $1/\epsilon \approx \log_2 \sqrt{N}$ (But it should be specially calculated for every dataset). Under this assumption, the vectors can be scaled up by $scale=1/\epsilon$ before quantizing them. When there is no loss of precision, the inner product will be scaled up by $1/\epsilon^2$ and the similarity results keep their order. Scaling up makes the data more scattered in $[-1,1]$ and can deduce the length error introduced by the quantization. We test our quantization approach on 128-d and 1000-d random data and SIFT1M dataset. The results show that after the quantization, the new vector length is about 5-10\% larger than the scaled vector, and the expansion of length on different quantized vectors is about 5\% of their average length, making the error tolerable.

\subsubsection{Angle Error} As the quantized vectors are no longer in the direction of the original ones, when we use the new vectors to calculate the cosine value in \ref{calcsim}, the value is, actually no way the same with the real value. This is the angle error. We may accept it when the change of angle is small as searching for the top $K$ nearest neighbors: cosine function has small derivative value when function value is around 1, so the cosine value will not be influenced much when the two vectors are intrinsically point to similar directions. For example, we find that when $N=128$, the angle between primitive and quantized vectors can be as much as 30 degrees, seems to be unacceptable. Using the scaling up trick proposed for length error, we can decrease the angle error from the quantization as well. The common case for the difference between the angle of $x, y$ and the angle of $x', y'$ after scaling up is just 5-10 degrees and less. Experiments on a higher dimension enhance this find. This angle error will finally add about 5\% relative error, on the top $K$ nearest vectors, when $K\leq 1000$, according to our trials.

In the discussion above, we used a conservative $scale$ parameter that is no larger than any components of all vectors. In most real-world problem, we can relax this restriction, and make it no larger than 98\% of all components, or even less. Experimental results show that a properly expanded $scale$ can accelerate calculation, holding the error acceptable.

\subsection{Control Selection Error by Extra Distance}
\label{sec:extra_dis}
Methods proposed in Section \ref{sec:err} can already help find similar vectors. Next we hope to further reduce the impact of these errors and make the result precise. 

First, we show how errors influences the distance we get. \ref{eq:inner_product:2} tells the relationship between the similarity ($\bm{X}\bm{Y}$) and the distance ($\bar{\bm{X}}\otimes\bar{\bm{Y}}$) we calculate. Especially when we quantize query with 4 bits and quantize document with 3 bits, the relationship is 
$$Distance = (constant) - 64ScaledSimilarity.$$

There is a linear relationship between the distance and the similarity. If the error is considered, then the scaled similarity (SS) can be written as
\begin{equation*}
\begin{split}
SS = (similarity + error_{angle})* (scale+ error_{query\ length})\\
* (scale + error_{doc\ length})
\end{split}
\end{equation*}
When we fix the query and find the top $K$ similar documents, the error on query length is fixed, and only the length error on documents and the angle error is floating. The length error round within 10\% of the scale, and the angle error can be assumed to be no larger than $\pm0.1$ for top documents. In total, the fluctuation range of the distance is no larger than 15\% of the whole range of the distance.

\subsubsection{Extra distance method}
\par
\begin{figure}[!htbp] 
    \centering 
    \includegraphics[width=0.45\textwidth]{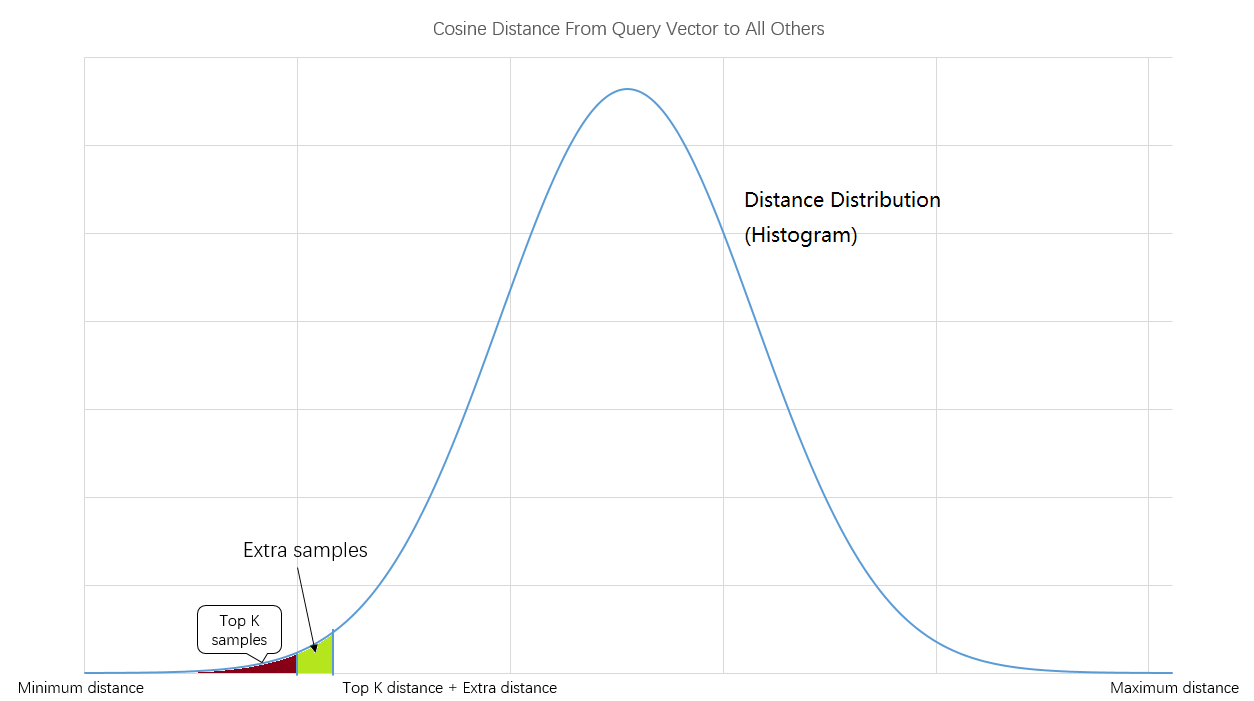}
    \caption{Sketch plot about the usage of extra distance. The gray area includes top $K$ similar vectors to the query, and the green area includes the following vectors with distances no larger than top $K$ distance plus extra distance. Vectors in both areas are sent to refine selection.} 
    \label{fig:extra_dis} 
\end{figure}

 Our strategy of improving search accuracy is as following: Given a query, distances from the query to candidate documents will be calculated by the quantization algorithm and then the minimum distance can be found to retrieve top $K$ similar documents (Gray part in Figure \ref{fig:extra_dis}). Besides, the minimum distance can optionally be added with an "extra distance" to prevent missing good results caused by the error. The documents whose distances are not greater than the minimum distance are recalled (Gray and green part in Figure \ref{fig:extra_dis}), and then the floating-point similarities between these documents and the query are calculated. Based on these similarities, the top $K$ results are finally returned.

\subsubsection{Choose the extra distance}

\par
\begin{figure}[!htbp] 
    \centering 
    \includegraphics[width=0.45\textwidth]{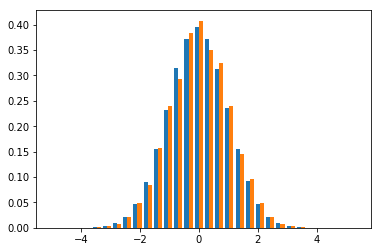}
    \caption{A sample of the distributions of exact similarity and quantized similarity after normalization. Practically they have the same distribution, but the components are arranged in totally different ways.} 
    \label{fig:sim_dist} 
\end{figure}

 In numerical experiments, we fixed one vector (as the query) and randomly generate massive vectors to calculate the pair-wise similarity by both real value and our quantization method. We normalize both similarities respectively so that results for both methods have 0 mean and 1 variance. A sample of the distributions after normalization is shown in Figure \ref{fig:sim_dist}. Notice that they are close to each other in distribution. This suggests that the boundaries of true top $K$ and the calculated top $K$ are close after normalization. 

\par Assume that the entire similarity space is divided into several buckets, then a vector with high similarity may drop down to no larger than a few buckets at a high probability. If we find the calculated top $K$ bounds and then consider the vectors in those lower buckets, and sort the original similarities of these vectors, the resulting top $K$ will have a high accuracy/recall rate.

\par As analyzed above, the fluctuation range of the distance is no larger than 15\% of the whole range of the distance. This theoretical upper bound can help to determine the upper bound for extra distance. In application with 4 bits * 3 bits quantization, we believe considering 5\% to 10\% of SS range is enough for high recall rates ($>99\%$).

\section{XFBQ based K-NN Search Algorithm}

\begin{algorithm}[!htbp]
  \caption{Fast K-NN Search Algorithm in CUDA}\label{alg:select}
    \algnewcommand{\LeftComment}[1]{\State \(\triangleright\) #1}
    \algblockdefx{ParallelFor}{EndParallelFor}
      [1]{\textbf{parallel in CUDA for} #1 \textbf{do}}
      {\textbf{end for}}
    
    \begin{algorithmic}[1]
      \Function{K-Select}{$[\bm{X}_1, \cdots, \bm{X}_n], \bm{Q}, K, extraDistance$}

        \LeftComment Quantize all document vectors and the query vector (with a scale factor)
        \State $[\bm{\bar{X}}_1, \cdots, \bm{\bar{X}}_n], \bm{\bar{Q}} = \Call{Quantize}{[\bm{X}_1, \cdots, \bm{X}_n], \bm{Q}}$
        \State \LeftComment Calculate distances between documents and the query
        \State $D = [d_1, \cdots, d_n]$ \Comment initialize distances
        \ParallelFor {$i \gets 1 : n$}
          \State $d_i = \Call{CalcDistance}{\bm{\bar{X}}_i, \bm{\bar{Q}}}$ 
        \EndParallelFor
        \State \LeftComment Histogram distances and select candidates by the $K$th minimum distance
        \State $H \gets [h_1, \cdots, h_M]$ \Comment initialize bins of the histogram
        \ParallelFor {$i \gets 1 : n$}
          \State \Call{Add}{$h_{d_i}$, 1}
        \EndParallelFor
        \State $maxDistance \gets (\arg\min_j\sum_{i=1}^j h_i \geq K) + extraDistance$ \Comment fix the boundary of top $K$ candidates
        \State $L \gets empty\ list$ \Comment initialize index array of candidates
        \ParallelFor{$index \gets 1 : n$}
          \If{$d_{index} \leq maxDistance$}
            \State Append $index$ to $L$
          \EndIf
        \EndParallelFor
        \State \LeftComment Refine candidates
        \State $\bm{S} \gets empty\ list$
        \ParallelFor{$i \gets 1 : L.length$}
          \State $similarity =$ \Call{CalcSimilarity}{$\bm{X}_{l_i}, \bm{Q}$}
          \State Append $(l_i, similarity)$ to $\bm{S}$
        \EndParallelFor
        \State $\bm{S} \gets$ \Call{SortBySimilarity}{$\bm{S}$} \Comment sort candidates by similarities
        \State \Return $[s_1, \cdots, s_{K}]$
      \EndFunction
    \end{algorithmic}
\end{algorithm}

As shown in Algorithm \ref{alg:select} , we perform a $K$-NN Search approach as following steps:

\begin{itemize} 
    \item Quantization. Quantize all document vectors and the query vector.
    \item Distance Calculation. Calculate distances between the quantized query vector and all quantized document vectors.
    \item Histogram and select. Histogram the distances and find the $K$th minimum distance as mentioned in Section \ref{sec:extra_dis}. Based on this distance, select the candidate documents.
    \item Refine. Sort candidate documents based on exact similarities calculated by original floating-point vectors. This step can also be done on CPU with little influence on overall performance , if GPU memory is limited.
\end{itemize}
\subsection{Details of the Selection Algorithm}
The details of these steps are described as follows.\\\\
\textbf{Quantization.}  The quantization of an N-dimension floating-point vector is optimized by the \textit{Advanced Vector Extensions} (AVX) intrinsic and some ingenious bit operations in CPU. The entire quantization process has two following steps. The first step is to quantize the floating-point vectors by \textit{Single Instruction Multiple Data} (SIMD) fashion since the quantization process for each dimension is independent. In the second step, the required bits are extracted from the quantized vectors and stored into \textit{uint64\_t} array for efficient access. By bitwise AND with elaborate bit masks and some other bitwise operations, 8-bit quantization can simultaneously be performed.\\\\
\textbf{Distance Calculation.}  In this step distances between the quantized query vector and all quantized document vectors are calculated on GPU. Before this, all quantized document vectors are reorganized as bundles of size 32 for warps - units of execution in GPU, and rearranged to column-based access pattern. With this structure of quantized document vectors, each warp can access the global memory in a high-efficient way. The calculation of the distance is implemented by the bitwise XOR, CUDA integer intrinsic \textit{\_\_popcll} and bitwise shift operations as shown in Equation \ref{eq:inner_product:6}. Distances are stored as integers of $\bar{\bm{X}} \otimes \bar{\bm{Y}}$, as they are equivalent to real inner products under a linear transformation.\\\\
\textbf{Histogram and Select.}  The top $K$ documents that are most similar to the query need to be selected with the information of obtained unordered distances. Traditional sequential method is to maintain a minimum heap and get top K results, but it is not totally friendly for parallel computing. Instead, a two-step parallel approach is used here. First, a histogram of the distances is parallelly built to find the $K$th minimum distance of all vectors. Since the integers stored by previous step take limited unique values, each unique value can take a bin in the histogram, and the process can be efficiently done. In the second step, with the $K$th minimum distance (often we add an extra distance as mentioned in Section \ref{sec:extra_dis}), the candidate top $K$ documents are selected from all documents. \\\\
\textbf{Refine.}  To improve the recall rate, the selected candidate results will be refined by exact similarities calculated from floating-point vectors. This process can be done on either CPU or GPU. For the GPU implementation, considering the feature of warp execution and 128-byte alignment of the L1 cache, vectors are grouped for calculation in a manner similar to loop unrolling. After calculation, candidate documents are sorted by exact similarities to get the top $K$ nearest neighbors.

\subsection{Complexity Analysis}

\subsubsection{Time Complexity}
In preprocessing step, we quantize all the document vectors from $N$ floating numbers into a space of $B_d$ bits $\times N$ on CPU and copy the quantized vectors to GPU. In practice we choose $B_d$ as 3. Time complexity here is $O(nN)$.

In searching step, we first calculate the distance between $n$ quantized document vectors and the query vector. Benefit from the new representation, we can take the special XOR operation between 64 pairs of tokens and add them up at the same time with $(B_d*B_q + B_d+B_q)$ bit operations and $(B_d*B_q-1)$ additions. Here $B_d$ and $B_q$ represent how many bits the document/query vectors are quantized into. Therefore, the time complexity here is still $O(\frac{B_dB_q}{64}nN)$. As $B_d$ and $B_q$ are fixed after preprocessing, this step can be seen as $O(nN)$ with a much smaller constant compared to brute force approach.

Next in histogram and select step, every distance is contributing to the corresponding bin in histogram on GPU parallelly, and the time cost does not exceed the size of the bin with the most distances -- In the worst case $O(n)$, but in average $O(n/(\text{Distance Scale}))$. Here Distance Scale is a number related to scale introduced in Section \ref{sec:err}. Generally speaking, time here can be ignored compared to distance calculation. Finding the boundary of top $K$ candidates takes $O({Distance Scale})$ time and finding top $K$ candidate vectors on GPU takes another $O(K)$ time. Finally in refine step as $O(K)$ candidates are taken, $O(KN+K\log K)$ time are needed for calculating the exact similarity and sort them.

In general, searching takes a time complexity of $O(\frac{B_dB_q}{64}nN)$. It can be at least 5 - 6 times faster than brute force way using the same computing resource.

\subsubsection{Space Complexity} 
Beside storing the vector information with $O(nN)$ space on CPU, we need an extra $O(\frac{B_d}{64}nN)$ space to store the quantized vectors on GPU. Distance result can be ignored compared to the vector storage. Detail of the storage has been shown in \ref{subsec:complexity}. Therefore, with the same space resources on GPU, our approach can deal with 10x data points compared to brute force way when data are all stored on GPU from the very beginning.

\section{Experiments Results}

In this section, we compare our approach with other popular methods on different public datasets. The baseline approach is a brute-force way of the similarity calculation implemented by ourselves that only taking advantage of the parallelization of GPUs. We also compared with the state-of-the-art method implemented by Faiss, including HNSW, IVF-HNSW on CPU, and IVF Flat,  IVF Product Quantization approach on GPU. All programs are run under Ubuntu 16.04 LTS with 20 Intel Core i7-6950X CPU @ 3.00GHz and 1 - 4 Nvidia Titan V GPU. As we did not optimize our program for multi-query requests, we represent results of single-query requests here. The metric we use for comparison is query per second (QPS) and precision at top 1/10/100/1,000. Here we define 
$$\text{Precision@i} = |\text{(Calculated top K)}\cap\text{(Real top K)}|/K.$$

We first take experiments on synthetic datasets to choose parameters $B_d$ and $B_q$ that can make the search fast and accurate. When $B_d$ and $B_q$ are no larger than 2, the precision of the result will be poor. $B_d, B_q \in {3,4}$ takes a balance between speed and accuracy. When $B_d=B_q=3$, assume sizeof(uint64) = 8 bytes and sizeof(float) = 4 bytes, by Table \ref{tab:complexity:0} we know XFBQ based inner product calculation only costs 28\% instructions and 9\% space compared to the brute force algorithm. When $B_d=3$ and $B_q=4$, 37\% instructions and 11\% space are cost compared to the brute force algorithm.

We choose two different open datasets for testing. Test purposes are different on these datasets. 

\begin{figure}[!htbp] 
    \centering 
    \includegraphics[width=0.45\textwidth]{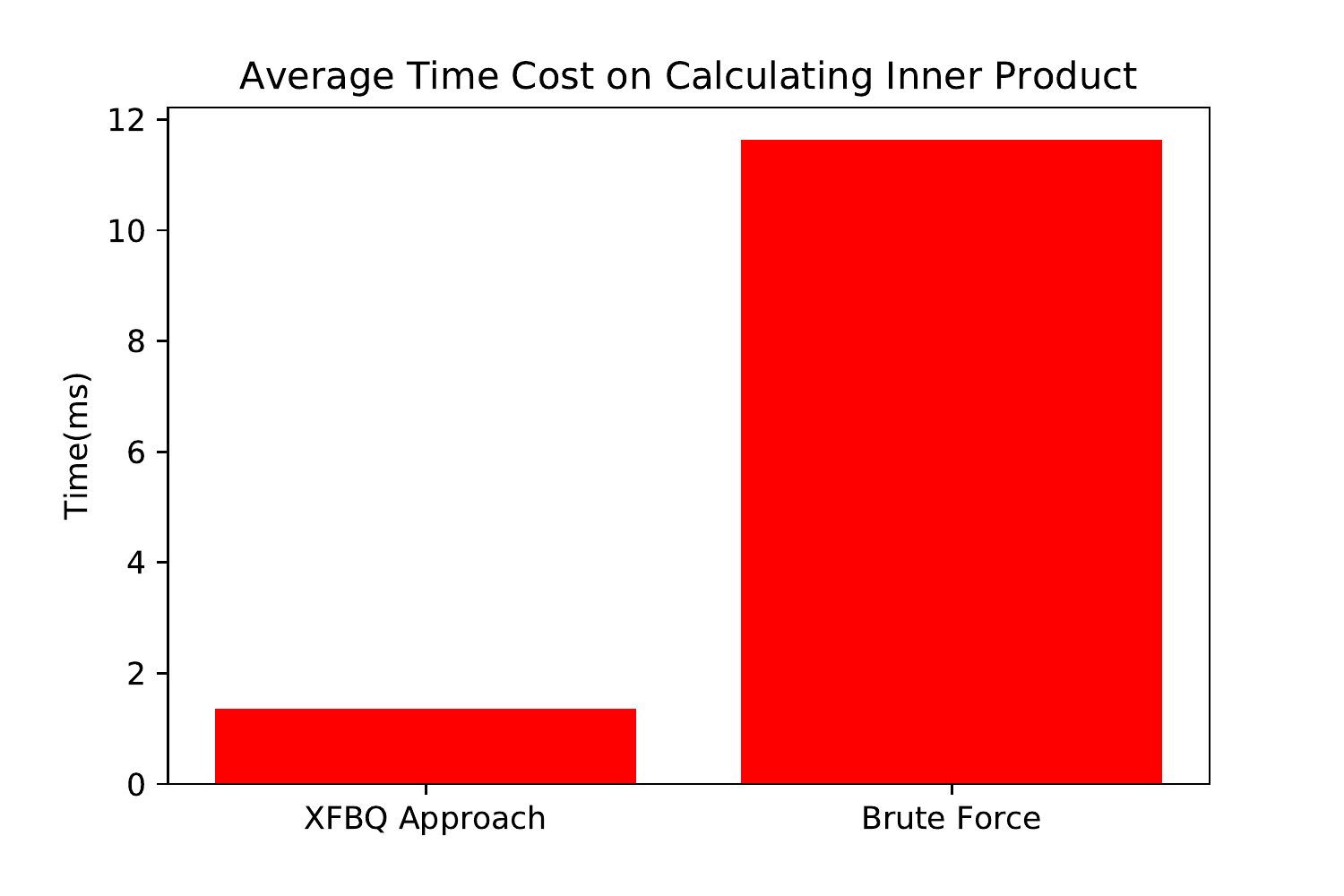} 
    \caption{Time cost on calculating inner product with and without XFBQ. XFBQ can save 90\% of the time cost.} 
    \label{fig:speed} 
\end{figure}

\begin{figure}[!htbp]
  \subfigure[Results on high precision part]{
    \begin{minipage}[b]{0.25\textwidth}
    \centering
    \includegraphics[width=\linewidth]{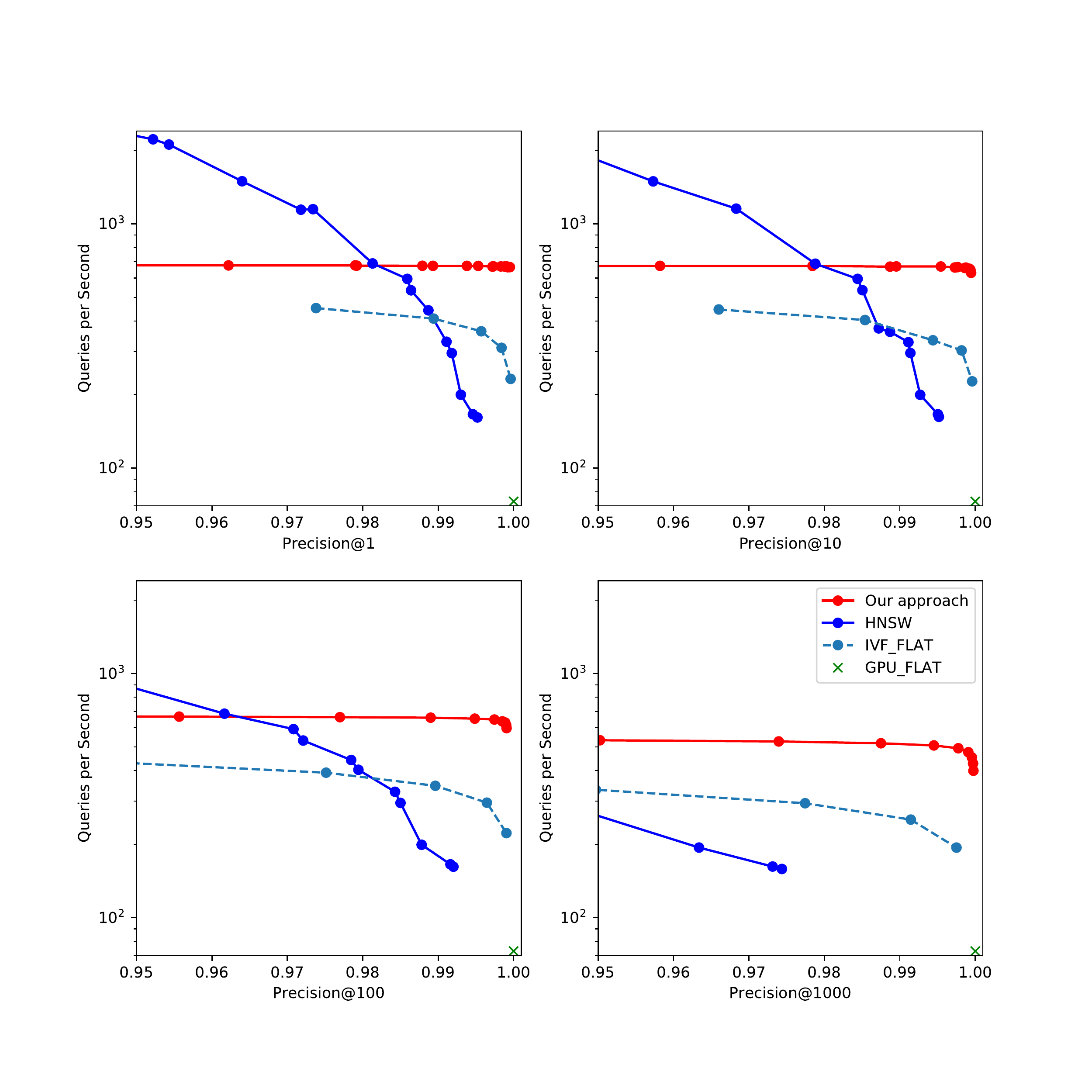}
    \label{fig:ailab:side:a}
    \end{minipage}
  }
  \subfigure[Results of IVF-PQ approach. Different trends for same label shows results of different PQ code length, 8/20/40 bytes from left to right.]{
    \begin{minipage}[b]{0.2\textwidth}
    \centering
    \includegraphics[width=\linewidth]{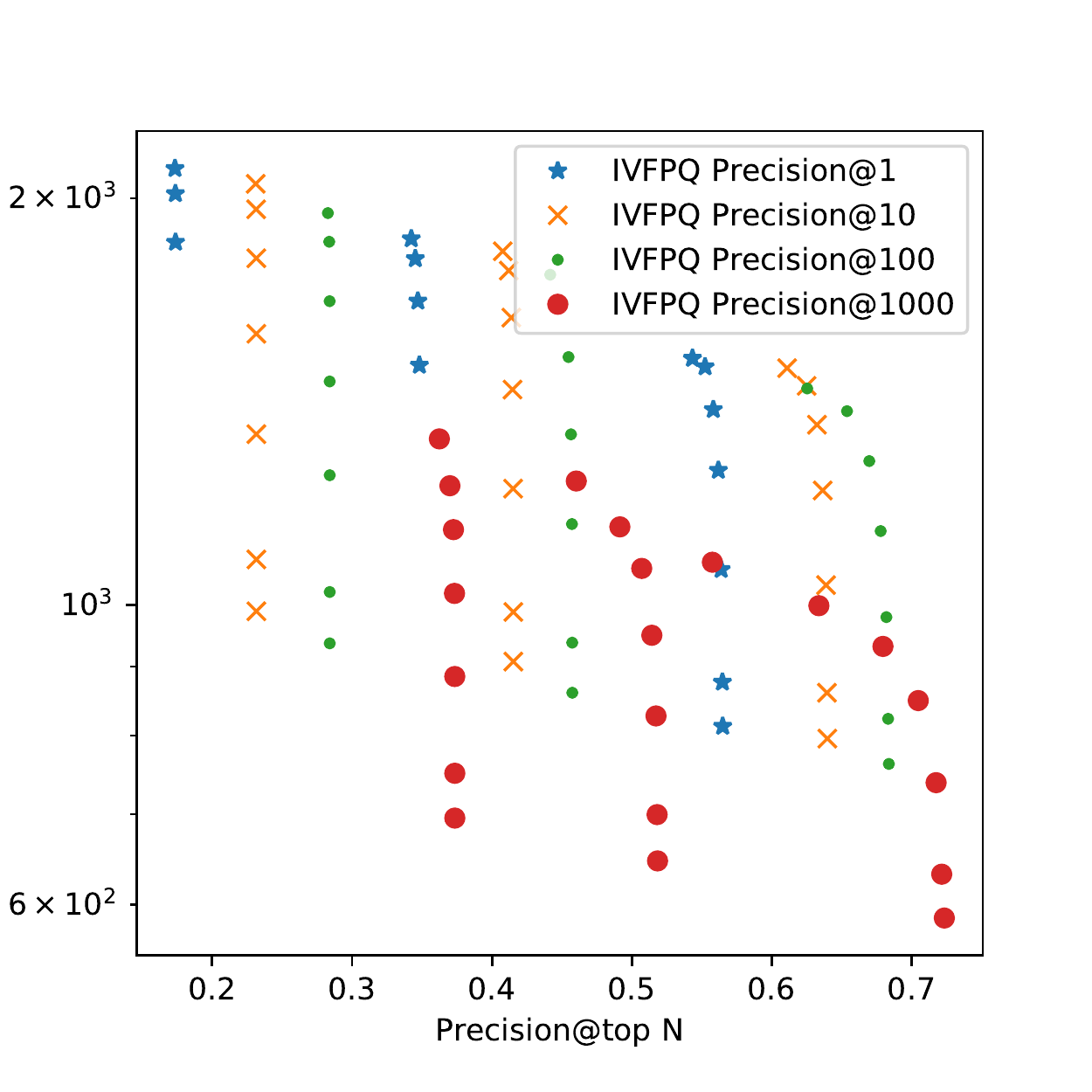}
    \label{fig:ailab:side:b}
    \end{minipage}
  }
    \caption[]{Queries per Second as functions of Precision at different top $K$ levels on Tencent AI Lab dataset. The IVF-PQ approach run by Faiss is on a separate figure due to low precision it reaches. Our approach based on binary quantization outperform all other state-of-the-art systems when a high precision is required. The efficiency is not influenced heavily when $K$ changes.}
    \label{fig:AILabResult}
\end{figure}

\textbf{Tencent AILab word embedding dataset} \cite{song2018directional} has 8,824,330 records for Chinese and English words. Each record is a 200-d vector. On this dataset we aim to test the speed of XFBQ based inner product calculation and the performance of our approach on single GPU. We randomly take 24,330 records in the dataset as queries and keep the others as dictionary. In our approach we choose $B_d=3, B_q=4$ as a common setting, and $scale=2.95$ based on the range of records. Precision are controlled by setting different extra distance (from 0 to 20). Figure \ref{fig:speed} show that our approach are on average 10 times faster than common parallel computing on inner product calculation. Results of the precision of k-NN search are shown in Figure \ref{fig:AILabResult}. IVF-PQ \cite{jegou2010product} provides a fast but low-precision search for nearest neighbors. It also costs great time on training. HNSW\cite{malkov2018efficient} also takes thousands of CPU seconds on training, and its efficiency drops down quickly when required $K$ increases.

\begin{table}[!htbp]
 \caption{Preprocessing Time on Tencent AI Lab Dataset}
  \centering
  \begin{tabular}{lcc}
    \toprule
    Method     & Resource Usage  & \parbox{2cm}{Preprocessing Time(s)} \\
    \midrule
    Our approach & 1 CPU + 1 GPU & 17 \\
    IVF3072FLAT &  1 CPU + 1 GPU & 18  \\
    IVF65536PQ8/20/40  & 1 CPU + 1 GPU & 440 - 800      \\
    HNSW20/40/50     & 20 CPU      & 480 - 600  \\
    \bottomrule
  \end{tabular}
  \label{tab:AILabTrain}
\end{table}
When high accuracy becomes a must, our approach has the highest QPS when using only one GPU. On the whole search process, we have 6 times faster than brute force way with optimized parallel implementation on GPU, and a 1.5x - 2x speed (measuring on same precision) compared to IVF3072 Flat search implemented by Faiss on GPU. Our approach also takes the least preprocessing time, which is similar to IVF Flat approach, as shown in Table \ref{tab:AILabTrain}. The space used by our approach is similar to HNSW. The result above shows that our approach can replace other state-of-the-art methods for k nearest neighbor search asking for high precision and a quick start.\\\\
\textbf{Deep100M} is the first 100 million vectors of dataset Deep1B \cite{babenko2016efficient}, which has 1 billion CNN representations for images with dimension 96. 
We design a test on only 100M records to avoid extreme long training time on those approaches for comparison. 
The query size is 10,000. We tested HNSW approach with 20 CPU cores. Other approaches are tested with 1 CPU and 1 or 4 GPUs in order to show the performance of our algorithm on multiple devices. In our approach we also choose $B_d=3$ and $B_q=4$, but set $scale=2.0$ to fit the data.

\begin{figure}[!htbp] 
    \centering 
    \includegraphics[width=0.45\textwidth]{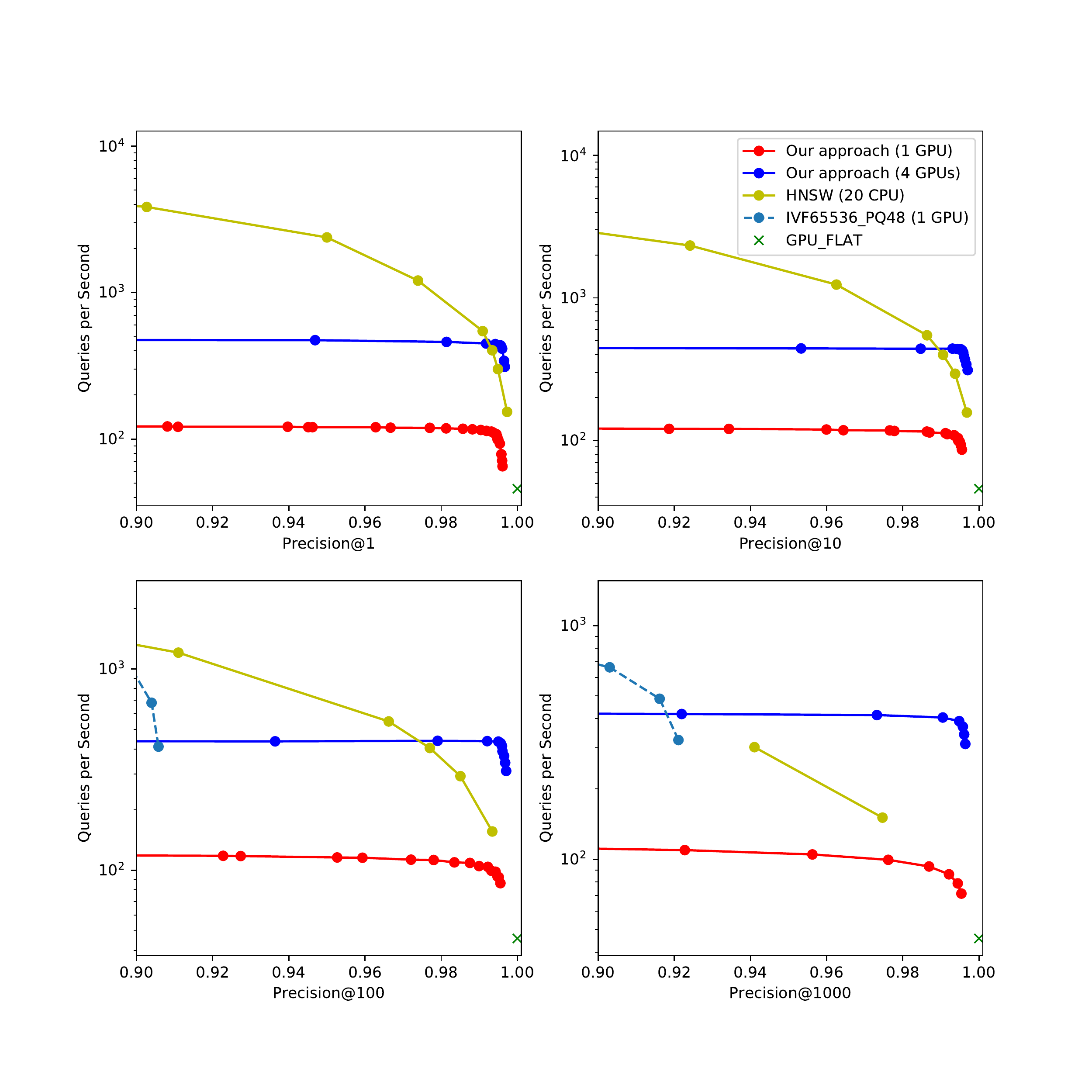} 
    \caption{Queries per Second as functions of Precision at different top $K$ levels on Deep100M dataset. Precision axis is rescaled with a lower limit of 0.9 to show the line for IVF-PQ method on some of the plot, as they cannot reach a high precision on all tasks. No green points are shown in the plot as that approach cannot reach such a high precision.}
    \label{fig:Deep100M} 
\end{figure}

The result is shown in Figure \ref{fig:Deep100M}. 
The scale of precision axis is different from that in Figure \ref{fig:ailab:side:a}, as we try to show the performance of the best among all IVF-PQ methods. 
The improvement on search efficiency shows that our approach fits well for multiple GPUs. When more devices are provided, the QPS increases linearly on our approach. In other words, our approach can be deployed on a distributed system. Our approach can deal with 2.5 billion tokens per gigabyte video memory, so a Nvidia P40 with 24GB memory can deal with 60 billion tokens, which exceeds the size of many real time searching problems. Our approach keeps a stable performance on all requests and is the fastest when precision is over 99\%, using 4 GPUs. Compared to HNSW, we also have much lower memory usage. 

\begin{table}[!htbp]
    \centering
    \caption{Preprocessing Time on Deep100M Dataset. For our approach, the time is used for encoding and copying to GPU. For others, here only record their time for training and adding index.} 
    \begin{tabular}{lcc} 
    \toprule
    Method     & Resource Usage  & \parbox{2cm}{Preprocessing Time(s)} \\
    \midrule
    Our approach & 1 CPU + 1 GPU  & 57 \\
    Our approach & 1 CPU + 4 GPU  & 20 \\
    FaissIVF65536PQ48  & 1 CPU + 1 GPU & 1424      \\
    HNSW32     & 20 CPU   & 6025  \\
    \bottomrule
    \end{tabular} 
    \label{table:Deep100M} 
\end{table}

Results on 1 GPU in Figure \ref{fig:Deep100M} reflects that our approach can still be dragged down by eager needs on computing power. This is the main drawback of it. As our approach only focus on how to calculate the similarity, it can be further improved by combining with other mature nearest neighbor search techniques focusing on reducing the searching area. For example, Locality Sensitive Hashing can be used in preprocessing and the search will then become non-exhaustive, taking away most of the calculation. In this way the searching will be n times faster than current speed when only one over n of the samples are chosen to be considered. Also, our techniques can be deployed on an inverted file system. Compared to IVF-PQ structure, we can provide higher precision with lower preprocessing time. Therefore, our approach has great potential on accelerating search speed.

\section{Conclusion}
In this work, we have presented a new approach for performing efficient k nearest-neighbor search using cosine similarity metric on GPU. We propose a new binary quantization method (XFBQ) for compressing calculation. This technique is combined with a special $k$-selection method using the calculated distance. Overall our techniques provide significant reductions on pre-searching time cost compared to other popular approximate nearest-neighbor search algorithms and keep a state-of-the-art searching efficiency, especially when high accuracy is needed. Since most of our work is on accelerating similarity calculation, our approach can be further combined with other popular techniques focusing on reducing search space, such as locality sensitive hashing, and inverted file system, to get a even faster speed. As a single high-performance GPU can handle the calculation work for hundreds of millions of vector productions, 
CPU resources can be liberated for works with higher complexity.


\bibliographystyle{ACM-Reference-Format}
\bibliography{references}

\appendix

\section{Proof of Proposition \ref{prop:limit}}
\label{sec:prof1}
\begin{proof}
First we prove  $|x-f_B(x)|\leq 2^{-B}$ $\forall x\in(-1,1), B\in\mathbb{N}^+$.
When $B=1$, by definition $x_1=0.5$ if $x\geq 0$. Otherwise $x_1=-0.5$. As $x\in(-1,1)$, in either situation $|x_1-x|\leq 2^{-1}$. The statement holds.

Now suppose the statement holds when $B=k-1$. We have $|x_{k-1}-x|\leq2^{-k+1}$. Then when $B=k$:

If $x>=x_{k-1}$, by definition we know $a_0=1$ in $f_k(x)$, and
$$
f_k(x)-f_{k-1}(x)=a_02^{-k} = 2^{-k}.
$$
\begin{align*}
\therefore x-f_k(x) &= x-f_{k-1}(x)+f_{k-1}(x)-f_k(x) \\
&= x-f_{k-1}(x)-2^{-k} \in [-2^{-k},2^{-k}].
\end{align*}
Thus $|x-f_k(x)|\leq 2^{-k}$. The statement also holds.

If $x<x_{k-1}$, we can prove the statement holds using the same technique. 

Therefore by deduction, $|x-f_B(x)|\leq 2^{-B}$ for any $B \in \mathbb{N}^+$.

The statement of convergence can be directly verified by the definition of uniform convergence using the result above.
\end{proof}





\end{document}